\documentclass[sn-mathphys]{sn-jnl}
\jyear{2022}%
\usepackage{mathtools}
\usepackage[utf8]{inputenc}
\usepackage{url}
\usepackage{moresize}
\usepackage{multirow}
\usepackage{rotating}
\usepackage{lscape}
\usepackage{setspace}
\renewcommand{\citet}{\cite}
\usepackage{graphicx}

\usepackage{hyperref}
\usepackage{algorithm}
\let\Algorithm\algorithm
\renewcommand\algorithm[1][]{\Algorithm[#1]\setstretch{1.1}}

\algrenewcommand{\algorithmiccomment}[1]{\hskip3px$\#$ #1}
\algdef{SE}[SUBALG]{Indent}{EndIndent}{}{\algorithmicend\ }%
\algtext*{Indent}
\algtext*{EndIndent}

\newcommand{\bb}{\textbf}

\begin{document}

\title{Combining Deep Learning with Good Old-Fashioned Machine Learning}

\author*[1]{\fnm{Moshe} \sur{Sipper}}\email{sipper@bgu.ac.il}

\affil*[1]{\orgdiv{Department of Computer Science}, \orgname{Ben-Gurion University}, \orgaddress{\city{Beer Sheva}, \postcode{84105}, \country{Israel}}}

\abstract{
We present a comprehensive, stacking-based framework for combining deep learning with good old-fashioned machine learning, called Deep GOld. Our framework involves ensemble selection from 51 retrained pretrained deep networks as first-level models, and 10 machine-learning algorithms as second-level models. Enabled by today's state-of-the-art software tools and hardware platforms, Deep GOld delivers consistent improvement when tested on four image-classification datasets: Fashion MNIST, CIFAR10, CIFAR100, and Tiny ImageNet. Of 120 experiments, in all but 10 Deep GOld improved the original networks' performance.
}

\keywords{machine learning, deep learning, image analysis, pattern recognition}

\maketitle

\section{Introduction}
\label{sec:intro}

The rapid rise of artificial intelligence (AI) in recent years has been accompanied (and enabled) by staggering advances both in software and hardware technologies.
Tools such as PyTorch \cite{paszke2019pytorch} for deep learning (DL), scikit-learn for machine learning (ML) \cite{scikit-learn,sklearn-website2022}, and graphics processing unit (GPU) hardware, all enable faster and better prototyping and deployment of AI systems than was possible a mere half-decade ago.

While deep learning has taken the world by storm, often---it would seem---at the expense of other computational paradigms, these (plentiful) latter are still quite alive and kicking. We propose herein to revisit stacking-based modeling \cite{wolpert1992stacked}, but within a comprehensive framework enabled by modern state-of-the-art software packages and hardware platforms.

As previously argued by \citet{Sipper2020,Sipper2021cml}, significant improvement can be attained by making use of models we are already in possession of anyway, through what they termed ``conservation machine learning'': conserve models across runs, users, and experiments--—and make use of all of them. Herein, focusing on image-classification tasks, we ask whether, given a (possibly haphazard) collection of deep neural networks (DNNs), can the tools at our disposal---specifically, ``good old-fashioned'' ML algorithms, many of which have been around for quite some time---help improve prediction accuracy. 

To wit, can we combine DL and ML in a manner that improves DL performance? We answer positively, with a major novelty being the use of most DL and ML models to date within a single, comprehensive framework. 

Section~\ref{sec:prev} discusses related previous work.
Section~\ref{sec:setup} describes Deep GOld---Deep Learning and Good Old-Fashioned Machine Learning---which employs 51 deep networks and 10 ML algorithms.
Section~\ref{sec:results} presents the results of 120 experiments over four image-classification datasets: Fashion MNIST, CIFAR10, CIFAR100, and Tiny ImageNet.
We end with a discussion in Section~\ref{sec:disc} and concluding remarks in Section~\ref{sec:conc}.

\section{Previous Work}
\label{sec:prev}

There are many works that involve some form or other of ensembling several models, and this section does not serve as a full review, but focuses on those papers found to be most relevant to our topic.

In an early work, \citet{OpitzShavlik96} presented a technique called Addemup that uses a genetic algorithm to search for an accurate and diverse set of trained networks. Addemup works by creating an initial population of networks, then evolving new ones, keeping the set of networks that are as accurate as possible while disagreeing with each other as much as possible. They tested these on three DNA datasets of about 1000 samples.

A few years later, \citet{zhou2002ensembling} presented an approach named GASEN (Genetic Algorithm
based Selective ENsemble) to select some neural networks from a pool of candidates, and assign weights to their contributions to the resultant ensemble. 
The networks had one hidden layer with five hidden units.
The efficacy of this method was shown for regression and classification problems over structural (non-image) datasets of a few thousand samples. Another work by \citet{west2005neural} studied financial decision applications, wherein a neural-network ensemble prediction was similarly reached by weighting the decision of each ensemble member.

A more recent example (one of many) of straightforward ensembling is given in \cite{berkhahn2019ensemble}, who presented an ensemble  neural-network model for real-time prediction of urban floods. Their ensemble approach used a number of artificial neural networks with identical topology, trained with different initial weights. The final result of maximum water level was the ensemble mean. Ensemble sizes examined were 1, 5, and 10.

In a similar vein, \citet{al2018virtual} trained multiple neural networks and combined their outputs using three combination strategies: simple average, weighted average, and what they termed a meta-learner, which applied a Bayesian regulation algorithm to the network outputs.
The application field considered was real-time production monitoring in the oil and gas industry, specifically, virtual flow meters that infer multiphase flow rates from ancillary measurements, and are attractive and cost-effective solutions to meet monitoring demands, reduce operational costs, and improve oil recovery efficiency.

\citet{kitamura2019ankle} trained five convolutional neural networks (CNNs) to detect ankle fractures in radiographic views. Model outputs were evaluated using both one and three radiographic views. Ensembles were created from a combination of CNNs after training. They implemented a simple voting method to consolidate the output from the three views and ensemble of models.

\citet{yan2018detecting} presented a malware detection method called MalNet, which uses a stacking ensemble with two deep neural networks---CNN and LSTM---as first-level learners, and logistic
regression as a second-level learner.

\citet{alzubi2019boosted} examined neural-network ensemble classification for lung cancer disease diagnosis.
They proposed an ensemble of Weight Optimized Neural Network with Maximum Likelihood Boosting (WONN-MLB), which essentially seeks to find optimal weights for a weighted (linear) majority vote.
\citet{ludwig2019applying} applied a neural-network ensemble to intrusion detection, again using weighted majority voting.

\citet{Shwartz2022} recently presented a cogent case for the use of XGBoost for tabular data, demonstrating that it outperformed deep models. They also showed that an ensemble comprising 4 deep models and XGBoost, predicting through weighted voting, worked best for the tabular datasets considered.

\citet{GHOSH2021104202} proposed an ensemble DNN for tumor detection in colorectal histology images. The mechanism consists of weights that are derived from individual models. The weights are assigned to the ensemble DNN based on their metrics and the ensemble model is then trained. The model is again re-trained by freezing all the layers, except for the fully connected and dense layers.

\citet{Paul2020} presented an ensemble DL method to detect retinal disorders. Their method comprised three pretrained architectures---DenseNet, VGG16, InceptionV3---and a fourth Custom CNN of their own design. The individual results obtained from the four architectures were then combined to form an ensemble network that yielded superb performance over a dataset of retinal images.

\citet{Elliott2020} examined Deep Q-learning, presenting an ensemble approach that improved stability during training, resulting in improved average performance. 

As noted above, \citet{Sipper2020,Sipper2021cml} presented conservation machine learning, which conserves models across runs, users, and experiments—and makes use of them. They showed that significant improvement could be attained by employing ML models already available anyway.

\section{Deep GOld: Algorithmic Setup}
\label{sec:setup}

Stacking (or Stacked Generalization) \cite{wolpert1992stacked} is an ensemble method that uses multiple models to tackle classification or regression problems. The main idea is to first train different models on the original problem. The outputs of these models are considered to be a first level, which are then passed on to a second level to perform the final prediction. 
The inputs to the second-level model are thus the outputs of the first-level models.

Our framework involves deep networks as first-level models and ML methods as second-level models. For the former we used PyTorch, one of the top-two most popular and performant deep-learning software tools \cite{paszke2019pytorch}. The module \texttt{torchvision.models} contains 59 deep-network models that were pretrained on the large-scale (over 1 million images), 1000-class ImageNet dataset \cite{ImageNet2009}. 

Of the the 59 models we retained 51 (8 models proved somewhat unwieldy or evoked a ``not implemented'' error). Each of the models was first retrained over the four datasets we experimented with in this paper: Fashion MNIST, CIFAR10, CIFAR100, and Tiny ImageNet.
As seen in Table~\ref{tab:datasets}, these datasets contain between 50,000 and 100,000 greyscale or color images in the training set, 10,000 images in the test set, with number of classes ranging between 10 -- 200. Retraining was necessary since the datasets contain images that differ in size and number of classes from ImageNet.

\begin{table}
\centering
\caption{Datasets.}
\label{tab:datasets}
\begin{tabular}{rcccl}
 Dataset & Images & Classes & Training & Test  \\ \hline
 Fashion MNIST & 28x28 grayscale & 10 & 60,000 & 10,000 \\
 CIFAR10 & 32x32 color & 10 & 50,000 & 10,000 \\
 CIFAR100 & 32x32 color & 100 & 50,000 & 10,000 \\
 Tiny ImageNet & 64x64 color & 200 & 100,000 & 10,000 \\
\end{tabular}
\end{table}

For retraining, we replaced the last FC (fully connected) 1000-class layer with a sequence of blocks comprising three layers: \{FC, batchnorm, leaky ReLU\}, denoted FBL. The final number of features of the original network was reduced to the dataset's number of classes through halving the number of nodes at each layer, starting with the closest power of 2. 
Consider en example: If the original network ended with 600 features, and the dataset contains 100 classes, then our modified network's final layers comprised a 512-node, 3-layer FBL block (512 being the closest power of 2 to 600), followed by a 256-node FBL, followed by a 128-node FBL, and ending with the 100 classes.
In addition, the first convolutional layer of the original network needed adjustment in some cases.
The retraining phase is detailed in Algorithm~\ref{alg:retrain}.

\begin{algorithm}
\small
\caption{Retrain 51 pretrained models}\label{alg:retrain}
\begin{algorithmic}[1]
\Statex
\Require
\Indent
\Statex \textit{dataset} $\gets$ dataset to be used
\Statex \textit{pretrained} $\gets$ \{alexnet, densenet121, densenet161, densenet169, densenet201, efficientnet\_b0, efficientnet\_b1, efficientnet\_b2, efficientnet\_b3, efficientnet\_b4, efficientnet\_b5, efficientnet\_b6, efficientnet\_b7, mnasnet0\_5, mnasnet1\_0, mobilenet\_v2, mobilenet\_v3\_large, mobilenet\_v3\_small, regnet\_x\_16gf, regnet\_x\_1\_6gf, regnet\_x\_32gf, regnet\_x\_3\_2gf, regnet\_x\_400mf, regnet\_x\_800mf, regnet\_x\_8gf, regnet\_y\_16gf, regnet\_y\_1\_6gf, regnet\_y\_32gf, regnet\_y\_3\_2gf, regnet\_y\_400mf, regnet\_y\_800mf, regnet\_y\_8gf, resnet101, resnet152, resnet18, resnet34, resnet50, resnext101\_32x8d, resnext50\_32x4d, shufflenet\_v2\_x0\_5, shufflenet\_v2\_x1\_0, vgg11, vgg11\_bn, vgg13, vgg13\_bn, vgg16, vgg16\_bn, vgg19, vgg19\_bn, wide\_resnet101\_2, wide\_resnet50\_2\}
\Comment{Networks pretrained over ImageNet dataset}
\EndIndent

\Ensure
\Indent
\Statex Retrained models and their test scores
\EndIndent

\vspace{4pt}

\State Load \textit{training set} and \textit{test set} of \textit{dataset}
\For{\textit{net} $\in$ \textit{pretrained}} 
    \State Replace \textit{net} final layer, and possibly adjust first convolutional layer
    \State Train entire \textit{net} for 20 epochs over \textit{training set} \Comment{mini-batch size: 64 (8 for Tiny ImageNet), optimizer: SGD}
    \State Save trained \textit{net} and \textit{test set} score 
\EndFor

\end{algorithmic}
\normalsize
\end{algorithm} 

Once Algorithm~\ref{alg:retrain} is run for all four datasets, we are in possession of 51 trained models per dataset. We can now proceed to perform the two-level prediction, as detailed in Algorithm~\ref{alg:ml}. 
Our interest herein was to study what one can do with models one has at hand.
Towards this end, we first selected from the 51 retrained models three random ensembles of networks, of sizes 3, 7, and 11. Each network of an ensemble was then run over both the training and test sets of the dataset in question (without any training---only feed-forward output computation). 
These first-level outputs were then concatenated to form an input dataset for the second level.
For example, if the ensemble contains 7 networks, and the dataset in question is CIFAR100, then the first level creates two datasets: a training set with 50,000 samples and 701 features, and a test set with 10,000 samples and 701 features (701: 7 networks $\times$ 100 classes + 1 target class).

\begin{algorithm}
\small
\caption{Two-level prediction}\label{alg:ml}
\begin{algorithmic}[1]
\Statex
\Require
\Indent
\Statex \textit{dataset} $\gets$ dataset to be used
\Statex \textit{ml\_algs} $\gets$ \{SGDClassifier, PassiveAggressiveClassifier, RidgeClassifier, LogisticRegression, KNeighborsClassifier, RandomForestClassifier, MLPClassifier, XGBClassifier, LGBMClassifier, CatBoostClassifier\}
\EndIndent

\Ensure
\Indent
\Statex Test scores for majority prediction, and for all ML algorithms
\EndIndent

\vspace{4pt}

\noindent\hskip-\leftmargin \Comment{level 1: generate datasets from outputs of retrained networks}
\For{ \textit{i} $\in$ \textit{\{3,7,11\}} }
    \State \textit{networks} $\gets$ pick $i$ networks at random from retrained networks of Algorithm~\ref{alg:retrain}
    \For{\textit{net} $\in$ \textit{networks}} 
        \State Run \textit{net} over \textit{training set} and \textit{test set}
        \State Accumulate generated outputs along with (known) targets
    \EndFor
    \State Generate 2 datasets, respectively: \textit{train-i}, \textit{test-i}
\EndFor

\vspace{4pt}

\noindent\hskip-\leftmargin \Comment{level 2: run ML algorithms over datasets generated by level 1}
\For{ \textit{i} $\in$ \textit{\{3,7,11\}} } 
    \For{\textit{alg} $\in$ \textit{ml\_algs}}
        \State Load \textit{train-i}, \textit{test-i}
        \State Run \textit{alg} to fit \textit{model} to \textit{train-i}
        \State Test fitted \textit{model} on \textit{test-i}
    \EndFor
\EndFor

\end{algorithmic}
\normalsize
\end{algorithm} 

After the first level produced output datasets, we passed these along to the second level, wherein we employed ten ML algorithms:

\begin{enumerate}
    \item \texttt{sklearn.linear\_model.SGDClassifier}: Linear classifiers with SGD training. 
    \item \texttt{sklearn.linear\_model.PassiveAggressiveClassifier}: Passive Aggressive Classifier \cite{crammer06}.
    \item \texttt{sklearn.linear\_model.RidgeClassifier}: Classifier using Ridge regression.
    \item \texttt{sklearn.linear\_model.LogisticRegression}: Logistic Regression classifier.
    \item \texttt{sklearn.neighbors.KNeighborsClassifier}: Classifier implementing the k-nearest neighbors vote.
    \item \texttt{sklearn.ensemble.RandomForestClassifier}: A random forest classifier.
    \item \texttt{sklearn.neural\_network.MLPClassifier}: Multi-layer Perceptron classifier, with 5 hidden layers of size 64 neurons each.
    \item \texttt{xgboost.XGBClassifier}: XGBoost classifier \cite{Chen:2016}. 
    \item \texttt{lightgbm.LGBMClassifier}: LightGBM classifier \cite{ke2017lightgbm}.
    \item \texttt{catboost.CatBoostClassifier}: CatBoost classifier \cite{catboost2017}.
\end{enumerate}

\section{Results}
\label{sec:results}

Unsurprisingly, we found significant differences in the runtime of the level-2 ML algorithms (Algorithm~\ref{alg:ml}). While some methods, such as RidgeClassifier and KNeighborsClassifier, were very fast, usually finishing within minutes, others proved slow (notably, XGBClassifier and CatBoostClassifier, which took several hours). 
While the number of samples of the generated ML datasets for the four problems studied is similar (identical to the original datasets---Table~\ref{tab:datasets}), the number of features differs by an order of magnitude:
with 10 classes for Fashion MNIST and CIFAR10, 100 classes for CIFAR100, and 200 classes for Tiny ImageNet, the latter two have 10 and 20 times more features than the former two, respectively. Some ML methods are known to scale less well with number of features.

ML runtimes for Fashion MNIST and CIFAR10 proved sufficiently fast to afford the use of hyperparamater tuning. Towards this end we used Optuna, a state-of-the-art, automatic, hyperparameter optimization software framework \cite{akiba2019optuna}, which we previously used successfully \cite{Sipper2021alacarte,Sipper2021addgBoost}. Optuna offers a define-by-run style user API where one can dynamically construct the search space, and an efficient sampling algorithm and pruning algorithm. 
Moreover, our experience has shown it to be fairly easy to set up. 
Optuna formulates the hyperparameter optimization problem as a process of minimizing or maximizing an objective function given a set of hyperparameters as an input. 
The hyperparameter ranges and sets are given in Table~\ref{tab:hyperparams}. With CIFAR100 and Tiny ImageNet we did not use Optuna, but rather ran the ML algorithms with their default values.

\begin{table}
\caption{Hyperparameter value ranges and sets used by Optuna.}
\label{tab:hyperparams}
\centering
\vspace{5pt}
\resizebox{0.99\textwidth}{!}{%
\begin{tabular}{r|c|l}
\textbf{Algorithm} & \textbf{Parameter} & \textbf{Values} \\ \hline
\texttt{SGDClassifier} & alpha & [1e-05, 1] \\
                       & penalty & \{`l2', `l1', `elasticnet'\} \\ \hline

\texttt{PassiveAggressiveClassifier} & C & [1e-02, 10] \\
                       & fit\_intercept & \{True, False\} \\
                       & shuffle & \{True, False\} \\ \hline

\texttt{RidgeClassifier} & alpha & [1e-3, 10] \\
                         & solver & \{`auto', `svd', `cholesky', `lsqr', `sparse\_cg', `sag', `saga'\} \\ \hline

\texttt{LogisticRegression} & penalty & \{`l1', `l2'\} \\
                            & solver & \{`liblinear', `saga'\} \\ \hline
                               
\texttt{KNeighborsClassifier} & weights & \{`uniform', `distance'\} \\
                              & algorithm & \{`auto', `ball\_tree', `kd\_tree', `brute' \} \\ 
                              & n\_neighbors & [2, 20] \\ \hline

\texttt{RandomForestClassifier} & n\_estimators & [10, 1000] \\
                       & min\_weight\_fraction\_leaf & [0, 0.5]  \\
                       & max\_features & \{`auto', `sqrt', `log2'\} \\ \hline

\texttt{MLPClassifier} & activation & \{`identity', `logistic', `tanh', `relu'\} \\ 
                       & solver & \{`lbfgs', `sgd', `adam'\} \\ 
                       & hidden\_layer\_sizes & \{(64,64), (64,64,64), (64,64,64,64), (64,64,64,64)\} \\ \hline

\texttt{XGBClassifier} & n\_estimators & [10, 1000] \\
                       & learning\_rate & [0.01, 0.2]  \\
                       & gamma & [0, 0.4]  \\ \hline

\texttt{LGBMClassifier} & n\_estimators & [10, 1000] \\
                       & learning\_rate & [0.01, 0.2]  \\
                       & bagging\_fraction & [0.5, 0.95]  \\ \hline

\texttt{CatBoostClassifier} & iterations & [2, 10] \\
                       & depth & [2, 10]  \\
                       & learning\_rate & [1e-2, 10]  \\ \hline
\end{tabular}
}
\end{table}

Table~\ref{tab:results} presents our results (we set a 10-hour limit on an ML algorithm's run of a row in the table, i.e., the level-2 loop of Algorithm~\ref{alg:ml}.)
A total of 120 experiments were performed: 4 datasets $\times$ ensembles of size 3, 7, and 11 $\times$ 10 complete runs per dataset.
In each experiment we generated level-1 datasets and then executed the ML algorithms, as delineated in Algorithm~\ref{alg:ml}.
We then compared three values:
1) the test score of the top network amongst the random ensemble (known from Algorithm~\ref{alg:retrain});
2) the test score of majority prediction, wherein the predicted class is determined through a majority vote amongst the ensemble's networks' outputs;
3) the test score of the top ML method.
The code is available at \url{https://github.com/moshesipper}.

\begin{landscape}
\begin{table}
\centering
\caption{Results for ensembles of 3, 7, and 11 random networks.
         Accuracy scores shown are over test sets. 
         Net: score of best network.
         Maj: score of majority prediction.
         ML: score of best ML method.
         For the latter, the ML method producing the best score is given in parentheses.
            RG: Classifier using Ridge regression;
            KN: k-nearest neighbors classifier;
            SG: Linear classifier with SGD training;
            PA: Passive Aggressive classifier;
            LR: Logistic regression;
            RF: Random Forest classifier;
            MP: Multi-layer perceptron;
            LG: LightGBM; 
            XG: XGBoost;
            CB: Catboost.
         }
\label{tab:results} 
\ssmall
\vspace{5pt}
\begin{tabular}{r|ccc|ccc|ccc}
Dataset & \multicolumn{3}{|c|}{3 networks} & \multicolumn{3}{c}{7 networks}  & \multicolumn{3}{|c}{11 networks}  \\ \hline
                & Net & Maj & ML & Net & Maj & ML & Net & Maj & ML \\

\hline
\multirow{10}{*}{Fashion-MNIST} & 91.97\% & 92.38\% & \bb{92.40\% (KN)} & 92.23\% & 93.12\% & \bb{93.38\% (KN)} & 94.22\% & 93.79\% & \bb{94.40\% (RG)} \\
 & 91.97\% & 92.05\% & \bb{92.50\% (KN)} & 92.40\% & \bb{93.26\%} & 93.25\% (KN) & 94.22\% & 93.62\% & \bb{94.57\% (KN)} \\
 & 92.32\% & \bb{92.95\%} & 92.80\% (KN) & 93.24\% & \bb{93.69\%} & 93.59\% (KN) & 93.24\% & 93.57\% & \bb{94.11\% (RG)} \\
 & 92.05\% & 92.39\% & \bb{92.61\% (KN)} & 94.01\% & 93.15\% & \bb{94.54\% (RG)} & 93.24\% & 93.71\% & \bb{93.92\% (KN)} \\
 & 91.67\% & 91.40\% & \bb{92.00\% (KN)} & 92.95\% & 93.57\% & \bb{93.95\% (KN)} & 94.01\% & 93.19\% & \bb{94.50\% (KN)} \\
 & 91.98\% & \bb{92.93\%} & 92.74\% (KN) & 92.27\% & 93.16\% & \bb{93.37\% (KN)} & \bb{93.86\%} & 93.75\% & 93.84\% (KN) \\
 & 92.14\% & 92.69\% & \bb{93.19\% (KN)} & 93.86\% & 93.24\% & \bb{94.00\% (RG)} & 93.63\% & 93.83\% & \bb{94.07\% (KN)} \\
 & 91.82\% & 92.27\% & \bb{92.51\% (RF)} & 92.27\% & 93.09\% & \bb{93.48\% (KN)} & 93.86\% & 93.81\% & \bb{94.25\% (KN)} \\
 & 93.86\% & 92.79\% & \bb{93.94\% (KN)} & 92.95\% & 93.14\% & \bb{93.82\% (RF)} & 94.01\% & 93.78\% & \bb{94.66\% (RG)} \\
 & 91.82\% & \bb{92.81\%} & 92.42\% (RG) & 93.86\% & 92.85\% & \bb{94.18\% (KN)} & 94.01\% & 93.53\% & \bb{94.39\% (RG)} \\
\hline
\multirow{10}{*}{CIFAR10} & 74.72\% & 71.00\% & \bb{75.87\% (KN)} & 75.60\% & 79.78\% & \bb{80.42\% (KN)} & 87.82\% & 80.94\% & \bb{88.29\% (RG)} \\
 & 71.67\% & 72.10\% & \bb{75.05\% (RG)} & 86.90\% & 80.54\% & \bb{87.53\% (RG)} & 83.57\% & 80.09\% & \bb{84.82\% (RG)} \\
 & 82.48\% & 82.03\% & \bb{83.30\% (KN)} & 87.33\% & 85.50\% & \bb{89.15\% (RG)} & 87.33\% & 80.82\% & \bb{89.12\% (RG)} \\
 & 74.82\% & 74.68\% & \bb{75.05\% (KN)} & 74.72\% & 75.76\% & \bb{79.29\% (KN)} & \bb{86.60\%} & 83.28\% & 86.43\% (RG) \\
 & 74.95\% & \bb{76.48\%} & 76.26\% (KN) & 86.90\% & 82.72\% & \bb{87.57\% (RG)} & 86.60\% & 79.93\% & \bb{87.24\% (RG)} \\
 & 75.60\% & 75.28\% & \bb{76.79\% (KN)} & 83.57\% & 81.31\% & \bb{83.95\% (KN)} & 87.22\% & 81.54\% & \bb{88.33\% (RF)} \\
 & 76.21\% & 77.77\% & \bb{78.27\% (KN)} & 74.00\% & 74.30\% & \bb{77.96\% (KN)} & \bb{86.90\%} & 84.44\% & 86.67\% (RG) \\
 & 86.90\% & 85.96\% & \bb{88.16\% (LR)} & \bb{86.90\%} & 84.74\% & 86.72\% (RG) & 86.60\% & 82.26\% & \bb{87.71\% (RG)} \\
 & 86.60\% & 81.76\% & \bb{86.92\% (KN)} & 86.90\% & 82.54\% & \bb{88.00\% (RG)} & 87.82\% & 83.36\% & \bb{89.56\% (RG)} \\
 & 76.43\% & 72.62\% & \bb{77.42\% (SG)} & 86.60\% & 77.97\% & \bb{87.08\% (RG)} & 87.82\% & 85.02\% & \bb{89.59\% (RG)} \\
\hline
\multirow{10}{*}{CIFAR100} & 48.86\% & 51.02\% & \bb{54.69\% (KN)} & 55.70\% & 55.11\% & \bb{59.50\% (RG)} & 60.76\% & 60.10\% & \bb{66.02\% (RG)} \\
 & 48.74\% & 44.95\% & \bb{51.68\% (KN)} & 60.76\% & 60.11\% & \bb{65.82\% (LR)} & 61.66\% & 62.50\% & \bb{67.30\% (RG)} \\
 & 48.74\% & 49.47\% & \bb{54.08\% (KN)} & 46.38\% & 51.97\% & \bb{54.61\% (KN)} & 9.58\% & 57.51\% & \bb{66.07\% (LR)} \\
 & 60.08\% & 55.30\% & \bb{63.67\% (SG)} & 61.30\% & 54.45\% & \bb{64.14\% (RG)} & 9.58\% & 58.43\% & \bb{64.86\% (RG)} \\
 & 47.06\% & 47.46\% & \bb{52.48\% (RG)} & 61.30\% & 57.22\% & \bb{64.08\% (RG)} & 60.08\% & 59.00\% & \bb{64.55\% (RG)} \\
 & 47.55\% & 50.94\% & \bb{53.64\% (RG)} & 60.08\% & 56.86\% & \bb{64.46\% (RG)} & 60.76\% & 56.07\% & \bb{64.25\% (LR)} \\
 & 46.55\% & 43.81\% & \bb{51.19\% (KN)} & 47.55\% & 47.91\% & \bb{52.79\% (SG)} & 9.58\% & 58.29\% & \bb{64.89\% (RG)} \\
 & 61.30\% & 57.47\% & \bb{63.86\% (RG)} & 48.86\% & 50.85\% & \bb{54.93\% (RG)} & 61.30\% & 57.09\% & \bb{65.07\% (RG)} \\
 & 46.30\% & 44.65\% & \bb{50.30\% (SG)} & 9.58\% & 53.56\% & \bb{56.81\% (KN)} & 9.58\% & 59.39\% & \bb{66.34\% (RG)} \\
 & 9.58\% & 33.88\% & \bb{44.23\% (SG)} & 61.66\% & 57.84\% & \bb{64.84\% (SG)} & 61.48\% & 58.78\% & \bb{65.20\% (LR)} \\
\hline
\multirow{10}{*}{Tiny ImageNet} & 55.77\% & 54.32\% & \bb{59.97\% (RG)} & 57.40\% & 56.06\% & \bb{63.32\% (RG)} & 67.30\% & 65.33\% & \bb{70.17\% (RG)} \\
 & 53.50\% & 54.83\% & \bb{59.61\% (LR)} & 54.83\% & 53.33\% & \bb{59.26\% (RG)} & 57.40\% & 63.72\% & \bb{66.00\% (RG)} \\
 & 58.34\% & 48.45\% & \bb{61.60\% (RG)} & 58.34\% & 60.59\% & \bb{64.30\% (RG)} & 57.23\% & 58.45\% & \bb{64.31\% (RG)} \\
 & 58.34\% & 59.91\% & \bb{63.05\% (RG)} & 55.77\% & 54.83\% & \bb{60.10\% (RG)} & 58.62\% & 60.42\% & \bb{65.19\% (RG)} \\
 & 53.50\% & 50.97\% & \bb{58.10\% (RG)} & 55.47\% & 53.11\% & \bb{60.11\% (RG)} & 56.20\% & 60.78\% & \bb{63.83\% (RG)} \\
 & 57.23\% & 47.54\% & \bb{59.88\% (RG)} & 58.62\% & 61.26\% & \bb{62.14\% (RG)} & 67.30\% & 64.33\% & \bb{69.88\% (RG)} \\
 & 58.62\% & 59.98\% & \bb{64.69\% (RG)} & 58.62\% & 62.41\% & \bb{66.24\% (RG)} & 56.20\% & 60.02\% & \bb{64.09\% (RG)} \\
 & 57.40\% & 56.14\% & \bb{62.76\% (LR)} & 56.16\% & 60.85\% & \bb{63.46\% (RG)} & 58.62\% & 62.44\% & \bb{65.94\% (RG)} \\
 & 54.76\% & 53.81\% & \bb{60.57\% (RG)} & 56.61\% & 57.70\% & \bb{63.76\% (RG)} & 58.34\% & 63.05\% & \bb{65.98\% (RG)} \\
 & 57.23\% & 48.52\% & \bb{60.31\% (RG)} & 58.34\% & 62.00\% & \bb{63.85\% (RG)} & 58.62\% & 62.90\% & \bb{66.46\% (RG)} \\
 
\end{tabular}
\normalsize
\end{table}
\end{landscape}

\section{Discussion}
\label{sec:disc}
As observed in Table~\ref{tab:results}, of the total of 120 experiments, an ML algorithm won in all but 10 experiments (4 were won by the retrained network, and 6 by majority prediction).

We note that classical algorithms, notably Ridge regression and k-nearest neighbors, worked best (they account for 104 of the wins).
They are also fast, scalable, and amenable to quick hyperparameter tuning. 
If one wishes to focus on a smaller batch of ML algorithms, these two seem like an excellent choice.

As noted in Section~\ref{sec:intro}, we often find ourselves in possession of a plethora of models, either collected by us through many experiments, or by others (witness our use of pretrained models herein). Benefiting from current state-of-the-art technology, Deep GOld leverages this wealth of models to attain better performance. One can of course tailor the framework to available deep networks and to a personal predilection for any ML algorithm(s).

\section{Concluding Remarks}
\label{sec:conc}
We presented Deep GOld, a comprehensive, stacking-based framework for combining deep learning with machine learning. Our framework involves ensemble selection from 51 retrained pretrained deep networks as first-level models, and 10 machine-learning algorithms as second-level models.
We demonstrated the unequivocal benefits of the approach over four image-classification datasets.

We suggest a number of paths for future research:
\begin{itemize}
    \item Further analysis of ML algorithms whose inputs are the outputs of deep networks. Do some ML methods inherently work better with such datasets?
    \item Currently, the features for level 2 comprise only the level-1 outputs. We might enhance this setup through automatic feature construction.
    \item Train (or retrain) the level-1 networks \textit{alongside} a level-2 ML model: 
    1) After each training epoch of the networks in the ensemble, generate a dataset from the network outputs;  
    2) a level-2 ML algorithm then fits a model to the level-1 dataset;
    3) the ML model generates class probabilities, which are used to ascribe loss values to the networks-in-training.
\end{itemize}

\section*{Acknowledgement}
I thank Raz Lapid for helpful discussions.

\section*{Compliance with Ethical Standards}
\begin{description}
\item[Disclosure of potential conflicts of interest:] M. Sipper declares that he has no conflict of interest. 

\item[Research involving human participants and/or animals:] This article does not contain any studies with human participants or animals performed by any of the authors.

\item[Informed consent:] N/A.
\end{description}

\bibliography{bib}

\end{document}